\documentclass{article}
\usepackage{spconf,amsmath,graphicx,bbm}
\usepackage{multirow, hhline, stmaryrd}

\DeclareMathOperator*{\argmin}{arg\,min}

\title{Attentive Adversarial Learning for Domain-Invariant Training}
\name{Zhong Meng, Jinyu Li, Yifan Gong}
\address{Microsoft Corporation, Redmond, WA, USA \\ \normalsize \{zhme, jinyl, ygong\}@microsoft.com}

\begin{document}
\ninept
\maketitle
\begin{abstract}

Adversarial domain-invariant training (ADIT) proves to be effective in
suppressing the effects of domain variability in acoustic
modeling and has led to improved performance in automatic speech
recognition (ASR). 
 In ADIT, an auxiliary domain classifier takes in \emph{equally-weighted} deep features from 
 a deep neural network (DNN) acoustic model and is trained to  improve their domain-invariance by optimizing an adversarial loss function. 
In this
work, we propose an attentive ADIT (AADIT) in which we advance the domain classifier with an attention mechanism to \emph{automatically weight} the input deep features according to their importance in domain classification. 
With this attentive re-weighting, 
ADDIT can focus on the domain
normalization of phonetic components that are more susceptible to domain
variability and generates deep features with improved domain-invariance
and senone-discriminativity over ADIT. Most importantly, the attention block serves only as an \emph{external} component to the DNN acoustic model and is not involved in ASR, so AADIT can be used to improve the acoustic modeling with any DNN architectures. More generally, the same methodology can improve any adversarial learning system with an auxiliary discriminator.
Evaluated on CHiME-3 dataset, the AADIT
achieves 13.6\% and 9.3\% relative WER improvements, respectively, over a multi-conditional model and a strong ADIT baseline. 

\end{abstract}
\section{Introduction}
\label{sec:intro}
The deep neural network (DNN) based acoustic models have been widely used
in automatic speech recognition (ASR) and have achieved extraordinary
performance improvement \cite{DNN4ASR-hinton2012,  yu2017recent, seide2011conversational}.
However, the performance of a multi-conditional acoustic model trained with
speech data from a variety of environments, speakers, microphone channels,
etc. is still affected by the spectral variations in each speech unit
caused by the inter-domain variability \cite{Li14overview}. Recently, adversarial learning
\cite{gan} has effectively improved the noise robustness of the DNN
acoustic model for ASR \cite{grl_shinohara, grl_serdyuk,
meng2018adversarial} or that of the deep embeddings for speaker verification \cite{meng2019asv} by using gradient reversal layer network
\cite{ganin2015unsupervised} or domain separation network \cite{dsn}.
Similar idea has also been applied to reduce the effect of inter-speaker
\cite{meng2018speaker, saon2017english}, inter-language
\cite{yi2018adversarial} and inter-dialect \cite{sun2018domain} variability
in acoustic modeling that is trained with speech from multiple speakers,
multiple dialects or multiple languages. We name all the above approaches
\emph{adversarial domain-invariant training (ADIT)} by referring to each
speaker, environment, language, etc. that contributes to the
condition-variability of the training data a \emph{domain}. 

To perform ADIT, an additional DNN domain classifier is introduced and is
jointly trained with the multi-conditional DNN acoustic model to
simultaneously optimize the primary task of minimizing the senone
classification loss and the secondary task of mini-maximizing the domain
classification loss. Through this adversarial multi-task learning
procedure, a shared feature extractor is learned as the bottom layers of the DNN acoustic model that maps the input speech frames from different domains into \emph{domain-invariant} and senone-discriminative deep hidden
features, so that further senone classification is based on representations
with the domain factor already normalized out.  After ADIT, only the DNN acoustic model is used to generate word transcription for test data from unseen domain through \emph{one-pass online} decoding \cite{meng2018speaker}. The domain classifier is not used in ASR. 
In this work, we focus on improving the domain classifier in ADIT to generate deep features with increased domain-invariance \emph{without} changing the DNN acoustic model.

In ADIT, the sequence of deep features generated by the feature
extractor are weighted equally before taken as the input to the domain
classifier. In fact, the deep features corresponding to different phonetic
components are affected nonuniformly by domain variability and show
different domain-discriminativity to the domain classifier. 
To improve ADIT, we introduce an
\emph{attention mechanism} to allow the domain classifier to attend to different
positions in time of deep feature sequence with \emph{nonuniform} weights. The
weights are automatically and dynamically determined by the attention
mechanism according to the importance of the deep features in domain
classification. We call this method \emph{attentive adversarial DIT
(AADIT)}. With AADIT, the domain classifier 
induces attentive reversal
gradients that emphasize on the domain normalization of more
domain-discriminative deep features, improving the domain invariance of the
acoustic model and thus the ASR performance over ADIT.

Self-attention \cite{vaswani2017attention} is a new technique to improve encoder-decoder end-to-end models \cite{cho2014learning, chorowski2015attention, chan2016listen}. It is used in \cite{zhang2018self} to enhance the performance of the generative adversarial network \cite{gan} by generating images based on cues from all feature locations. 
To the best of our knowledge, we introduce, for the first time, the attention mechanism only as an \emph{auxiliary} component to the \emph{external} of a DNN acoustic model to reduce the domain variability and improve ASR performance. Note that, similar to the domain classifier, the auxiliary attention block does not participate in the ASR decoding, so the proposed AADIT framework can be widely applied to acoustic model with any DNN architectures. AADIT can also improve the robustness in knowledge transferring for T/S learning \cite{li2014learning, meng2019conditional} as in \cite{meng2018adversarial}.
More generally, the same methodology can be used to enhance the capability of the discriminators in any generative adversarial network \cite{gan} or gradient reversal layer network \cite{ganin2015unsupervised} for improved domain \cite{ganin2015unsupervised, dsn_meng} and speaker \cite{meng2019asa} adaptation, speech enhancement \cite{pascual2017segan, meng2018cycle, meng2018afm}, speech synthesis \cite{kaneko2017generative, hsu2018disentangling}, voice conversion \cite{hsu2017voice}, speaker verification \cite{meng2019asv}, image generation \cite{gan, radford2015unsupervised} and translation \cite{isola2017image}, etc.

We perform AADIT to suppress speaker and environment variabilities of the DNN acoustic model and thus to improve ASR. We explore two types of local attention mechanisms: additive
attention and dot-product attention for AADIT and investigate the effect of
attention window size, key/query dimension, positional encoding and
multi-head attention on the ASR performance. Evaluated on CHiME-3 dataset,
AADIT of DNN acoustic model achieves 13.6\% and 9.3\% relative word error rate (WER) improvements over the
multi-conditional model and ADIT, respectively. 

\section{Adversarial Domain-Invariant Training}
\label{sec:adit}

ADIT aims at reducing the variances of hidden and output unit
distributions of the DNN acoustic model that are caused by the inherent
inter-domain variability in the speech signal. To achieve
domain-robustness, one solution is to learn a \emph{domain-invariant} and
\emph{senone-discriminative} deep hidden feature in the DNN acoustic model
through adversarial multi-task learning and make senone posterior
predictions based on the learned deep feature.  In order to do so, we need
a sequence of speech frames $\mathbf{X}=\{\mathbf{x}_{1}, \ldots,
\mathbf{x}_{T}\}, \mathbf{x}_t \in \mathbbm{R}^{r_x}, t=1, \ldots, T$, a sequence of
senone labels $\mathbf{Y}=\{y_{1}, \ldots, y_{T}\}$, $y_t \in \mathbbm{R}$
aligned with $\mathbf{X}$ and a sequence
of domain labels $\mathbf{D}=\{d_{1}, \ldots, d_{T}\}, d_t \in \mathbbm{R}$
aligned with $\mathbf{X}$.  We view
the first few layers of the acoustic model as a feature extractor network
$M_f$ with parameters $\theta_f$ that maps input speech frames $X$ from
different domains to intermediate deep hidden features
$\mathbf{F}=\{\mathbf{f}_1, \ldots, \mathbf{f}_T\}, \mathbf{f}_t\in
\mathbbm{R}^{r_f}$ and the upper layers of the acoustic model as a senone
classifier $M_y$ with parameters $\theta_y$ that maps the deep
features $\mathbf{F}$ to the senone posteriors $p(s|\mathbf{f}_t; \theta_y), s\in \mathbbm{S}$
as follows:

\begin{align}
	M_y(\mathbf{f}_t) = M_y(M_f(\mathbf{x}_t)) = p(s | \mathbf{x}_t;
		\mathbf{\theta}_f,
	\mathbf{\theta}_y).
	\label{eqn:senone_classify}
\end{align}

We further introduce a domain classifier network $M_d$
which maps the deep features $\mathbf{F}$ to the domain posteriors $p(u |
\mathbf{f}_t; \mathbf{\theta}_d)$, $u \in \mathbbm{U}$ as follows:
\begin{align}
	M_d(M_f(\mathbf{x}_t)) & = p(u | \mathbf{x}_t; \mathbf{\theta}_f,
	\mathbf{\theta}_d),
	\label{eqn:domain_classify}
\end{align}
where $u$ is one domain in the set of all domains $\mathbbm{U}$. 
To make the deep features $\mathbf{F}$ domain-invariant, the distributions of $\mathbf{F}$
from different domains should be as close to each other as
possible. Therefore, we jointly train $M_f$ and $M_d$ with an
adversarial objective, in which $\mathbf{\theta}_f$ is adjusted to \emph{maximize}
the frame-level domain classification loss
$\mathcal{L}_{\text{domain}}$ while
$\mathbf{\theta}_d$ is adjusted
to \emph{minimize} $\mathcal{L}_{\text{domain}}$ below:
\begin{align}
	& \mathcal{L}_{\text{domain}}(\mathbf{\theta}_f, \mathbf{\theta}_d)
	= - \frac{1}{T} \sum_{t=1}^{T} \log
	p(d_t | \mathbf{f}_t; \mathbf{\theta}_d)\nonumber \\
	& \quad \quad \quad \quad \quad \quad = -\frac{1}{T} \sum_{t = 1}^{T} \sum_{u\in
		\mathbbm{U}} \mathbbm{1}[u =
		d_t] \log M_d(M_f(\mathbf{x}_t)), \label{eqn:loss_cond1}
\end{align}
where $\mathbbm{1}[\cdot]$ is the indicator function which equals to 1 if the condition in the squared bracket is satisfied and 0 otherwise.
This minimax competition will first increase the discriminativity of 
$M_d$ and the domain-invariance of the deep features generated by $M_f$, and
will eventually converge to the point where $M_f$ generates extremely
confusing deep features that $M_d$ is unable to distinguish.

At the same time, we want to make $\mathbf{F}$ senone-discriminative by
minimizing the cross-entropy senone classification loss between the predicted senone posteriors
and the senone labels below:
\begin{align}
	\mathcal{L}_{\text{senone}}(\theta_f, \theta_y) & = -\frac{1}{T}\sum_{t = 1}^T
	\log p(y_t|\mathbf{x}_t;\theta_f, \theta_y) \nonumber \\
	&=-\frac{1}{T}\sum_{t = 1}^T \sum_{s\in
		\mathbbm{S}} \mathbbm{1}[s =
	y_t] \log M_y(M_f(\mathbf{x}_t)).
	\label{eqn:loss_senone}
\end{align}

In ADIT, the acoustic model network and the condition classifier
network are trained to jointly optimize the primary task of senone
classification and the secondary task of
domain classification with an adversarial objective function. 

\section{Attentive Adversarial Domain-Invariant Training}
\label{sec:aadit}

In ADIT, the mini-maximization of the domain classification loss (Eq.
\eqref{eqn:loss_cond1}) plays an important role in normalizing the
intermediate deep feature $\mathbf{F}$ against different domains. However, the
domain classification loss is still computed from a sequence of
\emph{equally-weighted} deep features.
In fact, not all deep features are equally affected by domain
variability and provide identical domain-discriminative information to the
domain classifier. For example, deep features extracted from voiced frames
are more domain-discriminative than those from the silence; deep features
aligned with vowels are in general more susceptible to domain variability
than those with consonants. To address this problem, we introduce an \emph{attention
machanism} to dynamically and automatically adjust the weights for the deep
features in order to put more emphasis on the domain normalization of more
domain-discriminiative deep features and therefore enhance the overall
domain-invariance of the deep features. The acoustic model with such a domain-invariant deep feature is expected to achieve improved ASR performance over ADIT.

\begin{figure}[htpb!]
	\centering
	\includegraphics[width=0.8\columnwidth]{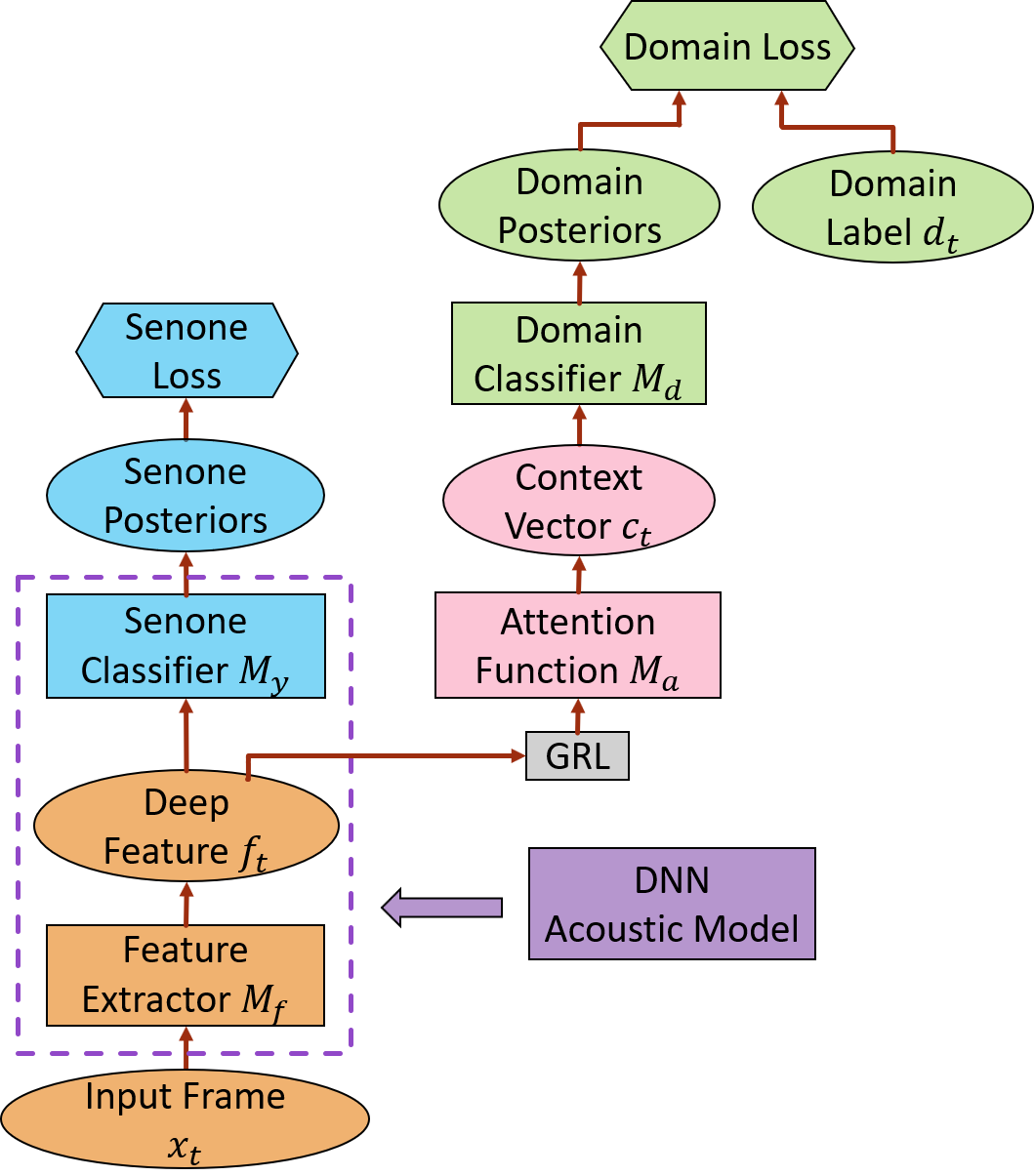}
	\caption{The framework of attentive adversarial domain-invariant training (AADIT) of the acoustic models. Only the DNN acoustic model consisting of $M_f$ and $M_y$ (on the left) are used for ASR on test data. $M_a$ and $M_d$ are discarded after AADIT.}
	\vspace{-6pt}
	\label{fig:aadit}
\end{figure}

In the proposed AADIT, we use soft local (time-restricted)
\cite{luong2015effective, povey2018time} self-attention
\cite{vaswani2017attention} because it is more suitable for ASR where the
input sequence consists of a relatively large number of speech frames. The
local attention we adopt selectively focuses on a small window of context
centered at the current time and can jointly attend different points in
time with different weights. Specifically, for each deep feature
$\mathbf{f}_t$ at
time $t$, the keys are the projections of deep features in an
$r_a$ dimensional space within the attention window of size $L + R + 1$, i.e., $\mathbf{K}_t =
\{\mathbf{k}_{t-L},
\ldots, \mathbf{k}_{t}, \ldots, \mathbf{k}_{t+R}\}$, and the query
$\mathbf{q}_t$ is the projection of $\mathbf{f}_t$ in the $r_a$ dimensional space, i.e., 
\begin{align}
	\mathbf{k}_t &= \mathbf{W}_k \mathbf{f}_t \label{eqn:key} \\
	\mathbf{q}_t &= \mathbf{W}_q \mathbf{f}_t, \label{eqn:query}
\end{align}
where $\mathbf{W}_k$ is a $r_a$ by $r_f$ key projection matrix, $\mathbf{W}_q$ is a $r_a$ by
$r_f$ query projection matrix and $L$ and $R$ are the length of left and
right context, respectively in the attention window.  The attention probability $\mathbf{a}_t$ of each current frame
$\mathbf{f}_t$ against all the context deep features in the attention window, i.e., $\mathbf{V}_t
= \{\mathbf{f}_{t-L}, \ldots, \mathbf{f}_{t}, \ldots, \mathbf{f}_{t+R}\}$, is computed by normalizing the
similarity scores $e_{t,\tau} \in \mathbbm{R}$ between the query
$\mathbf{q}_t$ and each key
$\mathbf{k}_{\tau}$ in the window $\mathbf{K}_t$, i.e.,
\begin{align}
	a_{t, \tau} = \frac{\exp(e_{t, \tau})}{\sum_{\tau' =
	t-L}^{t+R}\exp(e_{t,\tau'})},
	\label{eqn:attention_vector}
\end{align}
where $\tau = t-L,\ldots, t, \ldots, t+R$ and $a_{t,\tau} \in \mathbbm{R}$ is the $[\tau-(t-L)]^{\text{th}}$ dimension of
the attention probability vector $\mathbf{a}_t\in \mathbbm{R}^{L+R+1}$.
The similarity scores $e_{t,\tau}$ can be computed in two different ways according to
the type of attention mechanism applied:
\begin{itemize}
	\item Dot-product attention
	\begin{align}
		e_{t, \tau} = \frac{\mathbf{k}_{\tau}^{\top} \mathbf{q}_t}{\sqrt{r_a}},	
		\label{eqn:score_dot_product}
	\end{align}
	\vspace{-5pt}
	\item Additive attention
	\begin{align}
		e_{t, \tau} = \mathbf{g}^{\top} \tanh(\mathbf{k}_{\tau} +
		\mathbf{q}_t + \mathbf{b}),
		\label{eqn:score_additive}
	\end{align}
	where $\mathbf{g}\in \mathbbm{R}^{r_a}$ is a column vector, 
	$\mathbf{b}\in
	\mathbbm{R}^{r_a}$ is the bias column vector.
\end{itemize}
Therefore, a context vector $\mathbf{c}_t$ is formed at each time $t$ as a weighted sum of
the deep features in the attention window $\mathbf{V}_t$ with the attention vector
$\mathbf{a}_t$ serving as the combination
weights, i.e.,
	\begin{align}
		\mathbf{c}_{t} = \sum_{\tau = t-L}^{t+R} a_{t,\tau}
		\mathbf{f}_{\tau}.
		\label{eqn:context_vector}
	\end{align}
As shown in Fig.
\ref{fig:aadit}, we view the entire
attention process described in Eq.\eqref{eqn:key} to
Eq.\eqref{eqn:context_vector} as a single attention function $M_a(\cdot)$ with
parameters $\mathbf{\theta}_a = \{\mathbf{W}_k, \mathbf{W}_q, \mathbf{g}, \mathbf{b}\}$
which takes in the sequence of deep features $\mathbf{F}$ as the input and outputs
the sequence of context vectors $\mathbf{C} = \{\mathbf{c}_1,\ldots,
\mathbf{c}_T\}, \mathbf{c}_t\in \mathbbm{R}^{r_a}$, i.e., $\mathbf{c}_t =
M_a(\mathbf{f}_t)$.

The domain classifier $M_d$ then takes in the context vector $\mathbf{c}_t$
as the input to predict the frame-level domain posteriors for $u \in
\mathbbm{U}$.
\begin{align}
	p(u | \mathbf{x}_t; \mathbf{\theta}_f, \mathbf{\theta}_a, \mathbf{\theta}_d) &= M_d(\mathbf{c}_t) =
		M_d(M_a(\mathbf{f}_t)) \nonumber \\
	& = M_d(M_a(M_f(\mathbf{x}_t))).
	\label{eqn:domain_classify2}
\end{align}

To make the deep features $\mathbf{F}$ domain-invariant, $M_f$, $M_a$ and $M_d$ are jointly trained with an
adversarial objective, in which $\mathbf{\theta}_f$ is adjusted to \emph{maximize}
the frame-level domain classification loss
$\mathcal{L}_{\text{domain}}(\mathbf{\theta}_f, \mathbf{\theta}_a, \mathbf{\theta}_d)$ while
$\mathbf{\theta}_a$ and $\mathbf{\theta}_d$ are adjusted
to \emph{minimize} $\mathcal{L}_{\text{domain}}(\mathbf{\theta}_f,
	\mathbf{\theta}_a, \mathbf{\theta}_d)$ below:
\begin{align}
	&\mathcal{L}_{\text{domain}}(\mathbf{\theta}_f, \mathbf{\theta}_a, \mathbf{\theta}_d)
	 = -\frac{1}{T} \sum_{t=1}^{T} \log
	p(d_t | \mathbf{x}_t; \mathbf{\theta}_f, \mathbf{\theta}_a, \mathbf{\theta}_d)\nonumber \\
	& \qquad \quad = - \frac{1}{T}\sum_{t = 1}^{T} \sum_{u\in
		\mathbbm{U}} \mathbbm{1}[u =
		d_t] \log M_d(M_a(M_f(\mathbf{x}_t))). \label{eqn:loss_cond2}
\end{align}

In AADIT, the acoustic model network, the condition classifier network and
attention function are trained to jointly optimize the primary task of
senone classification and the secondary task of domain classification with
an adversarial objective function as follows 
\begin{align}
    (\hat{\mathbf{\theta}}_f, \hat{\mathbf{\theta}}_y) & = \argmin_{\mathbf{\theta}_y, \mathbf{\theta}_f}
	\mathcal{L}_{\text{senone}}(\mathbf{\theta}_f, \mathbf{\theta}_y) - 
	\lambda\mathcal{L}_{\text{domain}}(\mathbf{\theta}_f,
	\hat{\mathbf{\theta}}_a, \hat{\mathbf{\theta}}_d),
     \\
     (\hat{\mathbf{\theta}}_a, \hat{\mathbf{\theta}}_d) & =
     \argmin_{\mathbf{\theta}_a, \mathbf{\theta}_d}	\mathcal{L}_{\text{domain}}(\hat{\mathbf{\theta}}_f,
 	\mathbf{\theta}_a, \mathbf{\theta}_d),
\end{align}
where $\lambda$ controls the trade-off between $\mathcal{L}_{\text{senone}}$ and $\mathcal{L}_{\text{domain}}$, and $\hat{\mathbf{\theta}}_y, \hat{\mathbf{\theta}}_f, \hat{\mathbf{\theta}}_a$ and
$\hat{\mathbf{\theta}}_d$ are the optimized parameters.
 

The parameters are updated as follows via back propagation with stochastic gradient descent:
\begin{align}
	& \mathbf{\theta}_f \leftarrow \mathbf{\theta}_f - \mu \left[ \frac{\partial
		\mathcal{L}_{\text{senone}}}{\partial \mathbf{\theta}_f} - \lambda \frac{\partial
			\mathcal{L}_{\text{domain}}}{\partial
			\mathbf{\theta}_f}
		\right],
		\label{eqn:grad_f} \\
	& \mathbf{\theta}_a \leftarrow \mathbf{\theta}_a - \mu \frac{\partial
		\mathcal{L}_{\text{domain}}}{\partial \mathbf{\theta}_a},
		\label{eqn:grad_a} \\
	& \mathbf{\theta}_d \leftarrow \mathbf{\theta}_d - \mu \frac{\partial
		\mathcal{L}_{\text{domain}}}{\partial \mathbf{\theta}_d},
		\label{eqn:grad_s} \\
	& \mathbf{\theta}_y \leftarrow \mathbf{\theta}_y - \mu \frac{\partial
		\mathcal{L}_{\text{senone}}}{\partial \mathbf{\theta}_y},
	\label{eqn:grad_y}
\end{align}
where $\mu$ is the learning rate.
Note that the negative coefficient $-\lambda$ in Eq. \eqref{eqn:grad_f}
induces \emph{attentive} reversal gradient that maximizes $\mathcal{L}_{\text{domain}}$
in Eq.  \eqref{eqn:loss_cond1} 
to make the deep
feature domain-invariant. 
For easy implementation, a gradient reversal layer is introduced as in
\cite{ganin2015unsupervised}, which acts as an identity transform in the forward propagation
and multiplies the gradient by $-\lambda$ during the backward propagation.

Note that only the optimized DNN acoustic model consisting of $M_f$ and $M_y$ on the left side of  Fig. \ref{fig:aadit} is used for ASR on test data. The attention block $M_a$ and domain classifer $M_d$ (on the right) are discarded after AADIT.

We further extend the keys, queries and values with a one-hot encoding of the
relative positions versus the current time in an attention window as in \cite{povey2018time}
and compute the attention vectors based on the extended representations.
We also introduce a multi-head attention as in
\cite{vaswani2017attention} by projecting the deep features $H$
times to get $H$ keys and queries in $H$ different spaces. 
Note that the dimension of projection space for each attention head is
one $H^{\text{th}}$ of that in single-head attention to keep the number
of parameters unchanged.

\section{Experiments}
We perform AADIT of a multi-conditional acoustic model to suppress the speaker variability (AADIT-S) and environment variability (AADIT-E) for robust ASR.

\subsection{Baseline System}
As the baseline system, we first train a long short-term memory (LSTM)- hidden Markov model (HMM) acoustic model \cite{sak2014long,meng2017deep, erdogan2016multi} using
multi-conditional training data of CHiME-3 \cite{chime3_barker}. The
CHiME-3 dataset incorporates Wall Street Journal (WSJ) corpus sentences spoken under four challenging noisy environments, i.e, on buses, in cafes, in pedestrian areas, at street junctions and one clean environment, i.e., in booth, recorded using a 6-channel tablet.  The real far-field noisy speech from the 5th microphone channel in
CHiME-3 development data set is used for testing. A standard WSJ 5K word
3-gram language model is used for decoding.

We train the baseline LSTM acoustic model with 9137 clean and 9137 noisy
training utterances of CHiME-3 dataset by using cross-entropy criterion.
The 29-dimensional log Mel filterbank features together with 1st and 2nd
order delta features (totally 87-dimensional) for both the clean and noisy
utterances are extracted as in
\cite{li2012improving}. The features are fed as the input to the LSTM after
global mean and variance normalization. The LSTM has four 1024-dimensional hidden layers. 
Each hidden layer is followed by a 512-dimensional projection layer. 
The
output layer of the LSTM has 3012 output units corresponding to 3012 senone
labels. 
The multi-style LSTM
acoustic model achieves 19.23\% WER on the noisy test data.\footnote{Note that our
experimental setup does not achieve the state-of-the-art performance on
CHiME-3 (e.g., we did not perform beamforming, sequence training or
use recurrent neural network language model for decoding.) since our goal
is to simply verify the improved capability of AADIT in reducing inter-domain variability over ADIT.}

\vspace{-5pt}
\subsection{Adversarial Domain-Invariant Training}
\label{sec:exp_adit}
We further perform ADIT to reduce the speaker variability (ADIT-S) and environment variability (ADIT-E) of the
baseline multi-conditional LSTM with 9137 noisy training
utterances in CHiME-3. The $M_f$ is initialized with the
first $P$ layers of the LSTM and $M_y$ is initialized
with the rest $(7-P)$ hidden layers plus the output layer. $P$ indicates
the position of the deep hidden feature in the acoustic model. For ADIT-S,
the speaker classifier $M_d$ is a feedforward DNN with 3 hidden layers and
512 hidden units for each layer.  The output layer of $M_d$ has 87 units
predicting the posteriors of 87 speakers in the training set.  For ADIT-E,
$M_d$ is an environment classifier with the same architecture as in ADIT-S
except for the 5 output units predicting 5 
environments in CHiME-3.  $M_f$, $M_y$ and $M_d$ are jointly trained with an
adversarial multi-task objective as described in Section \ref{sec:adit}.
$P$ and $\lambda$ are fixed at $4$ and $0.5$ in our experiments. The ADIT-S
and ADIT-E LSTM acoustic models achieve 18.40\% and 18.31\% WER on the real
test data, respectively, which are 4.3\% and 4.8\% relative
improvements over the multi-conditional baseline.

\vspace{-5pt}
\subsection{Attentive Adversarial Domain-Invariant Training}
We further perform AADIT-S and AADIT-E with the same training data as ADIT. $M_f$, $M_y$ and $M_d$ have exactly the same
architectures as the ones for ADIT in Section \ref{sec:exp_adit} expect that $M_d$ has only one hidden layer to keep the number of parameters similar as that of ADIT. $P$ and $\lambda$ are also fixed at $4$ and $0.5$ for all the experiments.
\begin{table}[h]
\centering
\begin{tabular}[c]{c|c|c|c|c}
	\hline
	\hline
	$L+R+1$  & 15 & 21 & 25 & 31 \\
	\hline
	\multirow{1}{*}{\begin{tabular}{@{}c@{}} WER 
	\end{tabular}} & 17.89 & 17.63 & 17.89 & 18.07 \\
	\hline
	\hline
\end{tabular}
  \caption{The ASR WER (\%) of AADIT-S with addative attention of 
  LSTM acoustic models for different sizes of attention window ($L+R+1$) on real development set of CHiME-3.}
\label{table:context}
\end{table}

We first investigate the impact of attention window size $L+R+1$ on the ASR
WER via AADIT-S with additive attention in Table \ref{table:context}. In this work, we only use symmetric attention
window with $L=R$.
We fix the
key and query dimension $r_a$ at $512$. 
The WER begins to decrease when window size is larger than 21, so we choose $L=R=10$ for the following experiments.
Then we explore the effect of different key and query dimensions $r^a$ on
the ASR performance through AADIT-S with addative attention in Table \ref{table:key_dim}. 
The WER first decreases until reaching the
minimum at $r^a=512$ and then increases as $r^a$ grows larger.
Therefore, we fix $r^a$ at $512$ for the following experiments. Note that, by setting $r^a=512$, the number of learnable parameters are kept roughly the same as in ADIT.
\begin{table}[h]
\centering
\begin{tabular}[c]{c|c|c|c|c}
	\hline
	\hline
	$r^a$ & 256 & 512 & 1024 & 2048 \\
	\hline
	\multirow{1}{*}{\begin{tabular}{@{}c@{}} WER 
	\end{tabular}} & 18.53 & 17.63 & 18.01 & 18.19 \\
	\hline
	\hline
\end{tabular}
  \caption{The ASR WER (\%) of AADIT-S with addative attention of 
  LSTM acoustic models for different dimensions of keys and values
  ($r^a$) on real development set of CHiME-3.}
\label{table:key_dim}
\vspace{-6pt}
\end{table}

Further, we perform AADIT-S and AADIT-E with both additive and
dot-product attentions and summarize the WER results for different type of
domains in Table \ref{table:add_dot}. 
We see that AADIT-S
achieves 17.63\% WER with additive attention which is 8.3\% and 4.2\%
relatively improved over baseline multi-conditional LSTM and ADIT-S, respectively. AADIT-E performs significantly better
than AADIT-S with a WER of 16.61\% when using dot-product attention, which
is 13.6\% and 9.3\% relatively improved over  baseline
multi-conditional model and ADIT-E, respectively. Dot-product attention performs
similar to additive attention for AADIT. 

\begin{table}[h]
\centering
\begin{tabular}[c]{c|c|c|c}
	\hline
	\hline
	\multirow{2}{*}{\begin{tabular}{@{}c@{}} System 
		\end{tabular}} & \multirow{2}{*}{\begin{tabular}{@{}c@{}} Attention \\ Type \end{tabular}} & \multicolumn{2}{c}{Domain} \\
	\hhline{~~--}	
    & & \hspace{7pt} Speaker \hspace{7pt} & Environment\\
	\hline
	MC & - & 19.23 & 19.23 \\
	\hline
	ADIT & - & 18.40 & 18.31 \\
	\hline
	\multirow{2}{*}{\begin{tabular}{@{}c@{}} AADIT 
		\end{tabular}} &  AD & 17.63 & 16.82 \\
	\hhline{~---}
	&  DP & 17.67 & 16.61 \\
	\hline
	\multirow{2}{*}{\begin{tabular}{@{}c@{}} AADIT + PE 
		\end{tabular}} &  AD & 17.57 & 16.68 \\
	\hhline{~---}
	&  DP & 17.37 & 16.94 \\
	\hline
	\multirow{2}{*}{\begin{tabular}{@{}c@{}} MH AADIT 
		\end{tabular}} &  AD & 17.33 & 17.10 \\
	\hhline{~---}
	&  DP & 17.25 & 16.97 \\
	\hline
	\hline
\end{tabular}
  \caption{The ASR WERs (\%) of multi-conditional (MC) LSTM acoustic models, ADIT, single-head AADIT, single-head AADIT with positional encoding (PE) and multi-head (MH) AADIT on real development set of
  CHiME-3. Both the additive (AD) and dot-product (DP) attentions are used for each AADIT system. }
\label{table:add_dot}
\end{table}

We also investigate the effect of positional encoding on AADIT. In Table \ref{table:add_dot}, positional encoding does not consistently improve the AADIT, so we do not use it for the following experiments.
We further perform AADIT with multi-head additive and dot-product
attentions. The number of heads is fixed at $8$ and the key/query dimension
for each head is $512/8=64$. We observe that the multi-head AADIT-S only slightly improves the WER of single-head one, and multi-head AADIT-E does not further improve the WER. Considering the significantly better WER with less computational cost, we suggest using single-head AADIT-E for robust ASR.

\section{Conclusions}
We advance the domain classifier of ADIT with an attention mechanism to re-weight the deep features in a DNN acoustic model according to their importance in domain classification. With AADIT, the deep features more susceptible to domain variability are normalized with more emphasis and therefore, the overall domain-invariance of the acoustic model is greatly enhanced. The attention mechanism only serves as an auxiliary component to the external of the acoustic model that does not participate in ASR and thus can improve the DNN acoustic model with any architectures.

Evaluated on CHiME-3 dataset, the single-head AADIT achieves 13.6\% and 9.3\% relative WER gains over a multi-conditional LSTM acoustic model and a strong ADIT baseline, respectively. AADIT-E performs significantly better than AADIT-S. The additive and dot-product attentions achieve similar ASR performance. WERs of AADIT do not improve significantly with additional positional encoding and multi-head self-attention. 
\vfill\pagebreak

\bibliographystyle{IEEEbib}
\bibliography{refs}

\end{document}